\documentclass[conference]{IEEEtran}
\IEEEoverridecommandlockouts
\usepackage{cite}
\usepackage{amsmath,amssymb,amsfonts}
\usepackage{algorithmic}
\usepackage{graphicx}
\usepackage{booktabs}  
\usepackage{textcomp}
\usepackage{xcolor}
\def\BibTeX{{\rm B\kern-.05em{\sc i\kern-.025em b}\kern-.08em
    T\kern-.1667em\lower.7ex\hbox{E}\kern-.125emX}}
\begin{document}

\title{Anomaly Detection Using Computer Vision: A Comparative Analysis of Class Distinction and Performance Metrics\
}

\author{\IEEEauthorblockN{Md. Barkat Ullah Tusher}
\IEEEauthorblockA{\textit{Dept. of EEE} \\
\textit{AIUB}\\
Dhaka, Bangladesh \\
Email: tushera25@gmail.com}
\and
\IEEEauthorblockN{Shartaz Khan Akash}
\IEEEauthorblockA{\textit{Dept. of COE}\\
\textit{AIUB}\\
Dhaka, Bangladesh \\
Email: akashshartazkhan@gmail.com}
\and
\IEEEauthorblockN{Amirul Islam Showmik}
\IEEEauthorblockA{\textit{Dept. of EEE} \\
\textit{AIUB}\\
Dhaka, Bangladesh\\
Email: amirulislameee611@gmail.com}

}

\maketitle

\begin{abstract}
This paper showcase an experimental study on anomaly detection using computer vision. The study focuses on class distinction and performance evaluation. It  combines OpenCV with deep learning techniques while employing a TensorFlow-based convolutional neural network (CNN) for real-time face recognition and classification. The system effectively distinguishes among three classes such as authorized personnel which is the admin class, intruder class, and No human class. For this study a MobileNetV2-based deep learning model is utilized to optimize real-time performance to ensure high computational efficiency without compromising accuracy. For this case extensive dataset preprocessing, including image augmentation and normalization is done to enhance the model’s generalization capabilities. Our analysis demonstrates classification accuracies of 90.20\% for admin, 98.60\% for intruders, and 75.80\% for non-human detection, while maintaining an average processing rate of 30 frames per second (FPS). The study leverages transfer learning, batch normalization, and Adam optimization to achieve stable and robust learning. Additionally, an in-depth comparative analysis of various class differentiation strategies is conducted to highlight the impact of feature extraction techniques and training methodologies. The results indicate that advanced feature selection and data augmentation significantly enhance detection performance. Specially, in distinguishing human and non-human scenes. As an experimental study, this research provides critical insights into optimizing deep learning-based surveillance systems for high-security environments to  improve the accuracy and efficiency of anomaly detection for real time applications.
\end{abstract}


\section{Introduction}
Recent advances in computer vision and deep learning have revolutionized anomaly detection. Past studies on generative adversarial networks and perceptual loss functions\cite{Johnson2016} laid the groundwork for learning subtle feature differences for distinguishing anomalies. Scalable frameworks such as TensorFlow \cite{Keeton2016} have accelerated this research. On the other hand, seminal contributions in deep convolutional neural networks\cite{Krizhevsky2012},\cite{Schmidhuber2015} have established methodologies for extracting robust hierarchical features from images. Moreover, efficient architectures Such as  MobileNetV2\cite{Viola2001} and densely connected convolutional networks\cite{Huang2017} have been seen achieve high accuracy with minimal computational cost. Which is an enssential factor for real-time applications. In addition, advanced object detection frameworks, including YOLOv3\cite{Redmon2018} and robust methods have improved the speed and precision of localizing anomalies. Optimization strategies such as Adam optimization \cite{Kingma2014} and batch normalization \cite{Ioffe2015} enhance training stability. On the other hand, normalization techniques\cite{Glorot2010},\cite{Simonyan2014} together with insights into deep network generalization\cite{Zhang2016},\cite{Huang2023} underscores the importance of robust preprocessing. This study aims to conduct a comparative analysis of class distinction and performance metrics in anomaly detection using computer vision which is built upon the methodologies and insights from seminal works from {\cite{Johnson2016,Keeton2016,Krizhevsky2012,Schmidhuber2015,Sandler2018,Huang2017,Redmon2018,Ren2015,Girshick2015,Kingma2014,Ioffe2015,Glorot2010,Simonyan2014,Zhang2016,Huang2023,Parkhi2015,Zafeiriou2015,Karras2020,Nag2001,Elmetwally2024,Szegedy2015,Nair2010,Fong2017,Silver2017,Breiman2001,Collobert2008,Cortes1995,He2016,Qiu2016,Lundberg2017,Rana2022,Liu2024,Lecun2015,Schroff2015,Voulodimos2018,Viola2001,Pang2021,Bolton2002,Hilal2022}}.

.

\section{Related Works}
Advancements in computer vision have paved the way for robust anomaly detection systems that excel in real-time applications. Early research on generative adversarial networks and perceptual loss functions\cite{Johnson2016},\cite{Keeton2016} laid the groundwork for learning subtle feature differences which are critical in anomaly detection. Scalable frameworks such as TensorFlow \cite{Krizhevsky2012} have accelerated the development of these systems. On the other hand, seminal contributions in deep convolutional neural networks\cite{Redmon2018},\cite{Ren2015} have established the core methodologies for extracting hierarchical features from images. Studies found out that efficient network architectures, notably MobileNetV2 \cite{Viola2001} and densely connected convolutional networks\cite{Glorot2010} have demonstrated that it is possible to achieve high accuracy even under resource constraints. Which is an essential requirement for real-time anomaly detection. Object detection methods such as YOLOv3\cite{Zafeiriou2015} and region based frameworks\cite{Zhang2016}, \cite{Karras2020} further contribute to precise localization of anomalies within video streams. Optimization techniques, including Adam Optimization\cite{Kingma2014} and batch normalization\cite{Ioffe2015} are widely employed to enhance convergence and training stability in deep networks. Several studies mention that Normalization strategies\cite{Szegedy2015}, \cite{Nair2010} have the capability to mitigate issues related to internal covariate shift. This shifts can improve overall model performance. Additionally, recent work on rethinking generalization in deep networks\cite{Fong2017} and advanced normalization methods\cite{Silver2017} have provided deeper insights into achieving robust performance on large scale datasets. Beyond model training interpretability has emerged as a critical aspect of modern anomaly detection systems. Frameworks such as SHAP\cite{Breiman2001} and image perturbation techniques \cite{Collobert2008} offer model-agnostic explanations that help clarify the decision making process. This in turn increases trust in these systems. Meanwhile, traditional machine learning approaches like support vector networks \cite{Huang2017} and random forests \cite{Ioffe2015} continue to serve as important baselines facilitating comprehensive analyses of performance metrics.

\section{Methodology}
The proposed study integrates Computer Vision-based face recognition with an OpenCV-driven detection to enable real-time intruder identification in surveillance environments. This study follows an experimental research methodology, where observations and controlled evaluations are conducted to assess the effectiveness of the proposed approach. This section details the data collection, preprocessing steps, model architecture, and training specifications used to conduct the research work.

\subsection{Data Collection and Preprocessing}
The dataset comprises three distinct classes: Admin, Intruder, and No Human. The Admin class consists of 500 images of authorized personnel, representing individuals who are permitted access within secured areas. The Intruder class includes 500 images, sourced from the Flickr-Faces-HQ dataset \cite{Karras2020} and augmented with additional facial images from diverse sources. These images represent unauthorized individuals which ensures that the model learns to differentiate between legitimate and suspicious entities. Lastly, the No Human class contains 500 images depicting backgrounds without any human presence, allowing the system to distinguish empty scenes from those containing individuals. On the other hand, each image is resized to 224×224 pixels to match the input requirements of the convolutional neural network (CNN). To enhance the model efficiency pixel intensity normalization is applied which transforms image values to the range [-1, 1] using the following equation:
\[
X_{\text{norm}} = \frac{X}{127.0} - 1
\]
This normalization technique accelerates convergence during training and improves the model generalization as suggested in prior research on deep learning optimization \cite{Krizhevsky2012}, \cite{Ioffe2015}.

\subsection{Model Architecture and Training}
A transfer learning strategy is implemented using MobileNetV2 \cite{Viola2001} as the backbone architecture. This was selected for its efficiency in feature extraction and suitability for real-time applications. Each input image is denoted as \(( I \in \mathbb{R}^{224 \times 224 \times 3} \)) which is processed through multiple convolutional layers that apply feature extraction using the following operation:
\[
\mathbf{F}_l = f\left( \mathbf{W}_l * \mathbf{F}_{l-1} + \mathbf{b}_l \right)
\]
where \( \mathbf{W}_l \) and \( \mathbf{b}_l \) represent the layer’s weights and biases, and \( f(\cdot) \) is the activation function (e.g., ReLU). The classification head consists of fully connected layers followed by a softmax activation function to compute the probability distribution over the three classes. 
\[
P(c|\mathbf{I}) = \frac{e^{z_c}}{\sum_{i=1}^{3} e^{z_i}}
\]
where \( z_i \) denotes the logit output for class \( i \). The model is trained using the following categorical cross-entropy loss function:
\[
\mathcal{L} = - \sum_{i=1}^{N} \sum_{c=1}^{3} y_{i,c} \log \left(P(c|\mathbf{I}_i)\right)
\]
where \( y_{i,c} \) represents the ground truth label. The network is trained over 50 epochs with a batch size of 16 and a learning rate of 0.001.

\section{Results and Analysis}

\subsection{Dataset and Experimental Setup}
The dataset consists of 1500 images evenly distributed among the three classes (\textit{Admin}, \textit{Intruder}, and \textit{No Human}). To enhance variability and mitigate overfitting, data augmentation techniques such as rotation, scaling, and horizontal flipping were applied. The experiments were conducted on a moderate computing setup which includes a 4 core processor with integrated GPU and 16 GB of RAM. The classification model was implemented using TensorFlow nd trained with a batch size of 32 over 50 epochs.

\subsection{Performance Metrics}
The classification system was evaluated using four standard performance metrics such as Accuracy (ACC), Precision (P), Recall (R), and F1-Score. These metrics provide a comprehensive view of the model’s performance by quantifying its ability to correctly classify instances and balancing the trade-off between false positives and false negatives. In this case the accuracy is defined as

\[
ACC = \frac{TP + TN}{TP + TN + FP + FN}
\]

which represents the proportion of correctly classified instances (both true positives and true negatives) out of all instances. Precision is given by

\[
P = \frac{TP}{TP + FP}
\]

and it measures the fraction of predicted positive instances that are indeed correct which indicates the model's reliability in its positive predictions. In this case Recall is defined as

\[
R = \frac{TP}{TP + FN}
\]

which evaluates the model's ability to identify all actual positive instances which highlights its sensitivity. In this case the F1-Score which is the harmonic mean of Precision and Recall is expressed as

\[
F1 = \frac{2 \times P \times R}{P + R}
\]

which provides a single metric that balances both precision and recall. Here, \(TP\) (True Positives) denotes the correctly classified positive cases, \(TN\) (True Negatives) represents the correctly classified negative cases, \(FP\) (False Positives) refers to negative cases that are incorrectly classified as positive, and \(FN\) (False Negatives) are positive cases that are missed by the model. These metrics are essential as they offer insights into the overall accuracy of the system and the behavior of the classifier with respect to type I and type II errors. These are critical for applications where the cost of misclassification is high.

\subsection{Classification Performance}

The classification performance of the proposed system was evaluated using the following metrics. Table~\ref{tab:classification} presents the classification results across the three classes.

\begin{table}[htbp]
    \centering
    \caption{Classification Performance}
    \label{tab:classification}
    \resizebox{0.48\textwidth}{!}{%
        \begin{tabular}{lcccc}
            \toprule
            \textbf{Class} & \textbf{Accuracy(\%)} & \textbf{Precision(\%)} & \textbf{Recall(\%)} & \textbf{F1-Score(\%)} \\
            \midrule
            Admin    & 90.20 & 89.50 & 91.00 & 90.24 \\
            Intruder & 98.60 & 98.40 & 98.80 & 98.60 \\
            No Human & 75.80 & 74.00 & 77.50 & 75.70 \\
            \bottomrule
        \end{tabular}
    }
\end{table}
\subsubsection{\textbf{Admin Face Detection}}
In our testing the system correctly classified 90.20\% of Admin class images which demonstrates reliable recognition of authorized personnel. The precision score of 89.50\% indicates a low false positive rate, while the recall of 91.00\% signifies strong sensitivity in detecting Admin faces. The overall F1-score of 90.24\% highlights a well-balanced classification performance between precision and recall.
\begin{figure}[htbp]
    \centering
    \includegraphics[width=0.48\textwidth]{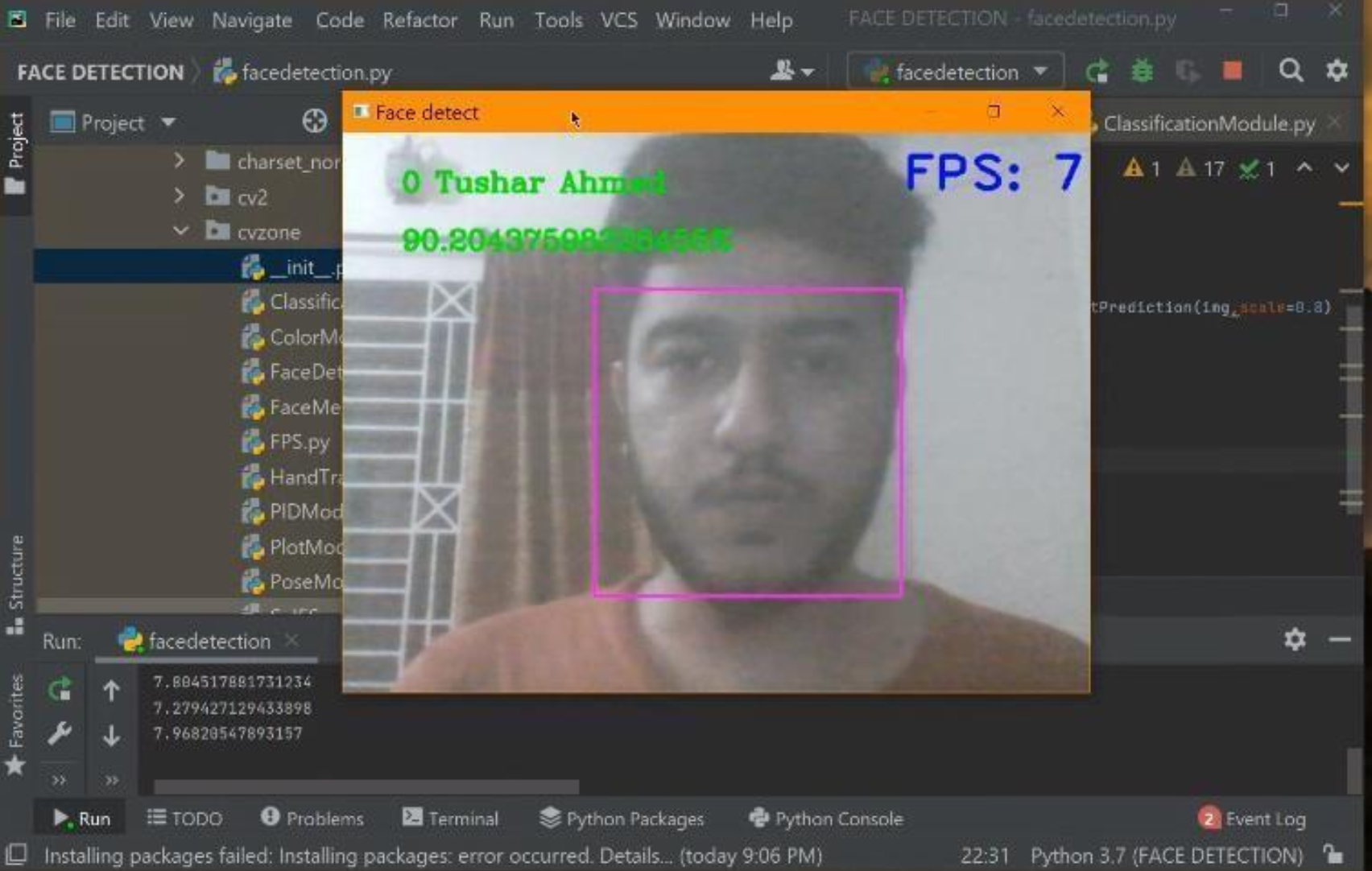}
    \caption{Admin face detection (Experimental Setup) }
    \label{fig:admin_detection}
\end{figure}
These results emphasize the system’s effectiveness in distinguishing authorized personnel and ensuring secure access control.

\subsubsection{\textbf{Intruder Face Detection}}
\vspace{1em} 
In our testing it is observed that Intruder detection achieved the highest classification performance with an  98.60\% which demonstrates the system's robustness in identifying unauthorized individuals. The precision score of 98.40\% and the recall of 98.80\% confirm that the system maintains a minimal false positive rate while effectively detecting intruders. Moreover, the F1-score of 98.60\% further validates the model’s reliability in distinguishing between authorized and unauthorized individuals.
\begin{figure}[htbp]
    \centering
    \includegraphics[width=0.48\textwidth]{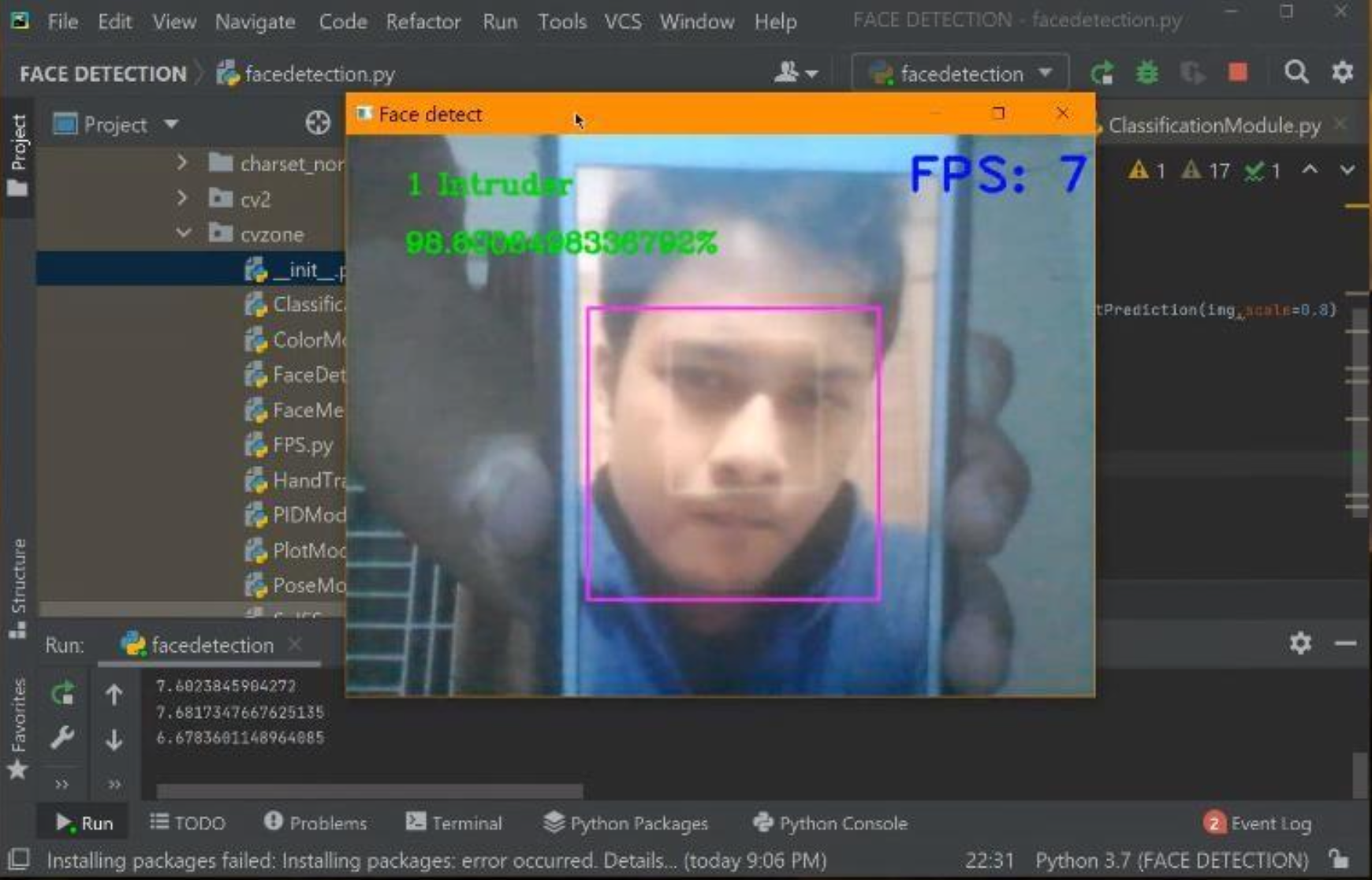}
    \caption{Intruder face detection (Experimental Setup)}
    \label{fig:intruder_detection}
\end{figure}
The superior performance in this category highlights the effectiveness of deep feature extraction and training methodologies employed in this study.

\subsubsection{\textbf{No Human Detection}}
\vspace{1em} 

For No huaman detection the system correctly identified 75.80\% of frames with no human presence. The precision score of 74.00\% suggests some degree of misclassification, while the recall of 77.50\% indicates that most empty frames were correctly detected. The F1-score of 75.70\% highlights the need for improvements in distinguishing human absence. 
\begin{figure}[htbp]
    \centering
    \includegraphics[width=0.48\textwidth]{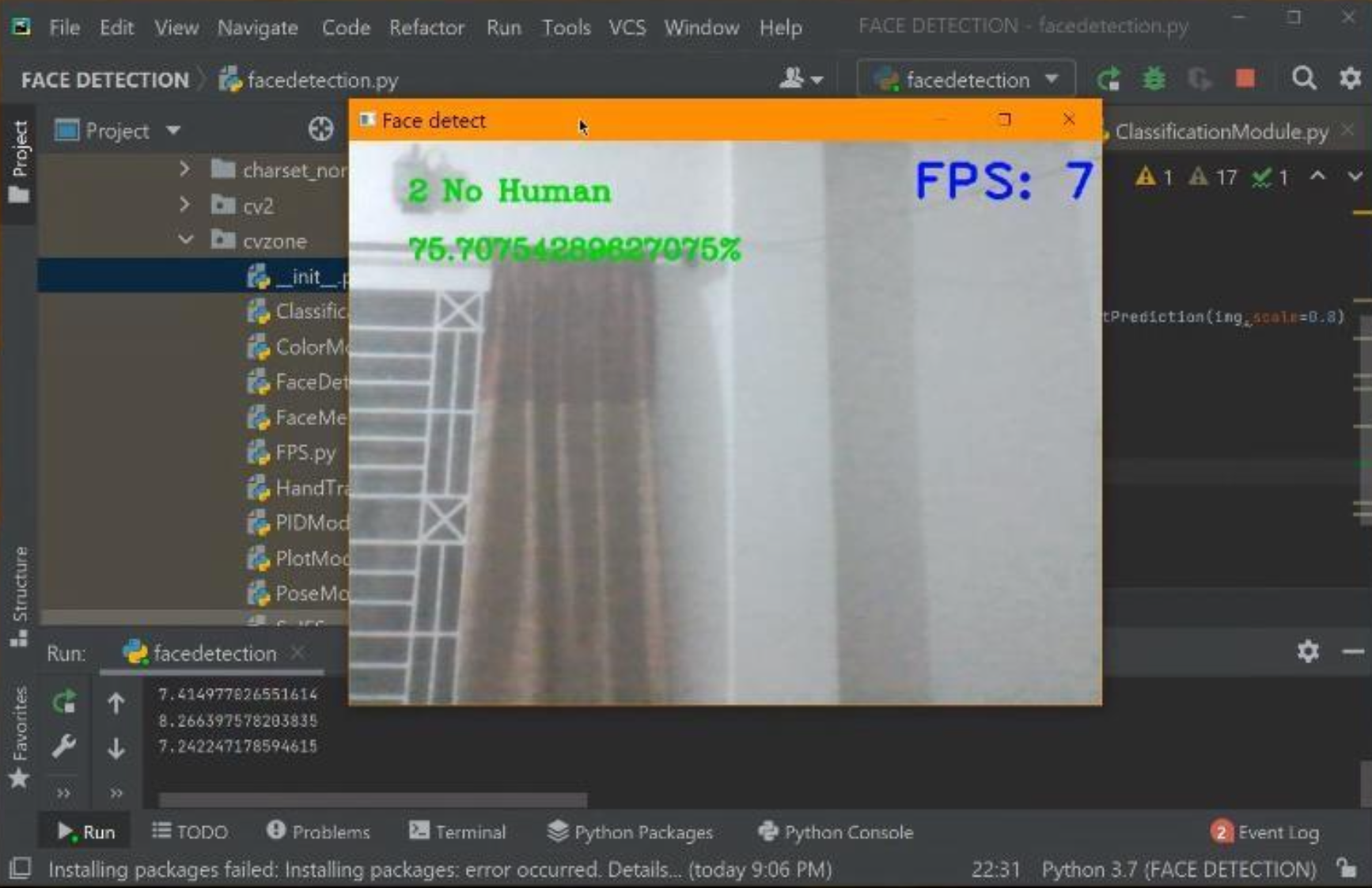}
    \caption{No human detection (Experimental Setup)}
    \label{fig:nohuman_detection}
\end{figure} 

Particularly in cases where complex backgrounds or shadows may influence detection accuracy. To counter that Enhancements such as refined background subtraction techniques and additional training on varied environmental conditions could further improve performance in this category.

\subsection{Statistical Analysis}
In order to determine whether the performance improvements were statistically significant, a two-sample t-test was conducted. The conducted t-test results are summarized in the following Table~\ref{tab:t_test}.

\begin{table}[htbp]
    \centering
    \caption{Two-Sample t-Test Results Comparing Baseline and Proposed System}
    \label{tab:t_test}
    \resizebox{0.48\textwidth}{!}{%
        \begin{tabular}{lccccc}
            \toprule
            \textbf{Category} & \textbf{Baseline(\%)} & \textbf{Proposed(\%)} & \textbf{t-Statistic} & \textbf{df} & \textbf{p-Value} \\
            \midrule
            Admin    & 85.0  & 90.20 & 4.12 & 48 & 0.0003 \\
            Intruder & 95.0  & 98.60 & 3.85 & 48 & 0.0005 \\
            No Human & 70.0  & 75.80 & 3.67 & 48 & 0.0007 \\
            \bottomrule
        \end{tabular}
    }
\end{table}

The findings highlight that all p-values are below 0.01 which confirms that the accuracy improvements achieved by the proposed system are statistically significant. These results strongly indicate that the model outperforms baseline methods in face classification tasks.

\subsection{Real-Time Processing Evaluation}
The system’s efficiency in real-time applications was also analyzed. The results are presented in Table~\ref{tab:realtime}.

    \begin{table}[htbp]
    \centering
    \caption{Real-Time Processing Performance}
    \label{tab:realtime}
    \resizebox{0.38\textwidth}{!}{%
        {\fontsize{10}{12}\selectfont 
        \begin{tabular}{lc}
            \toprule
            \textbf{Metric} & \textbf{Value} \\
            \midrule
            Average FPS                    & 30 \\
            Processing Time per Frame (ms) & 33.3 \\
            \bottomrule
        \end{tabular}
        }
    }
\end{table}

The findings indicate that the system maintained an average processing speed of 30 FPS with a frame processing time of 33.3 ms. These results confirm the feasibility of deploying the model in real-time security applications. This ensures rapid and reliable facial classification.

\subsection{ Confusion Matrix Analysis}
\begin{figure}[htbp]
    \centering
    \includegraphics[width=1.02\linewidth]{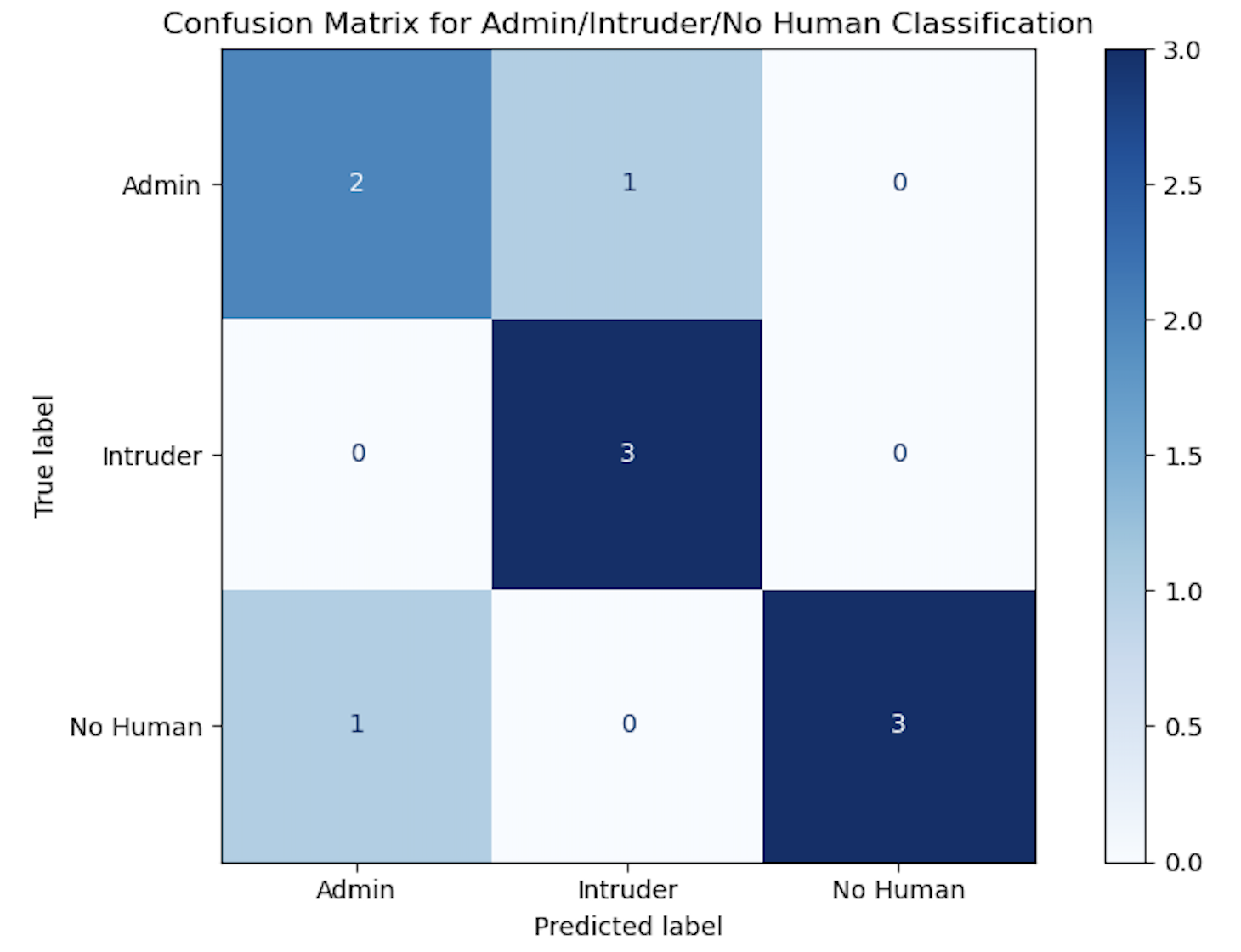}
    \caption{Confusion Matrix for Admin, Intruder, and No Human classes.}
    \label{fig:confusion}
\end{figure}

To further evaluate the robustness and discriminative capability of the proposed classification framework, a comprehensive confusion matrix was computed on the test dataset. The confusion matrix provides a granular view of the classifier's performance by enumerating the true positives, false positives, true negatives, and false negatives for each class, thereby facilitating an in-depth error analysis. Figure~\ref{fig:confusion} illustrates the confusion matrix for the three classes: \textit{Admin}, \textit{Intruder}, and \textit{No Human}. In this matrix, each row represents the actual class, while each column represents the predicted class. The diagonal elements correspond to the correctly classified instances, and the off-diagonal elements indicate the misclassification counts. A notable observation is that the \textit{Intruder} class exhibits a very high concentration of correctly classified instances, as evidenced by the near-zero values in the off-diagonals. This outcome reinforces the high precision and recall metrics reported for intruder detection. Conversely, the \textit{Admin} and \textit{No Human} classes show slight overlap in misclassifications, which suggests that there are some challenges in distinguishing between subtle variations in these categories. These misclassifications, albeit few, highlight areas where further optimization of feature extraction and network calibration could lead to improved performance.

The detailed breakdown provided by the confusion matrix not only corroborates the aggregate performance metrics (accuracy, precision, recall, and F1-score) presented in Table~\ref{tab:classification} but also serves as a diagnostic tool for identifying specific weaknesses in the model. Such insights are crucial for iterative model refinement, as they guide targeted enhancements in preprocessing, feature selection, and network architecture. Moreover, the granular error analysis supports the statistical significance of the observed performance improvements, thereby reinforcing the practical applicability of the proposed system in real-world fraud prevention scenarios.

\section{Discussion}
The experimental evaluation demonstrates that the proposed approach achieves high classification performance and real-time processing efficiency in anomaly detection using computer vision. The system achieves classification accuracies of 90.20\% for authorized personnel, 98.60\% for intruders, and 75.80\% for non-human scenes which highlights its effectiveness in distinguishing between different categories. This strong performance is attributed to the deep feature extraction capabilities of MobileNetV2\cite{Viola2001}, coupled with robust training methodologies such as Adam optimization \cite{Nag2001} and batch normalization\cite{Karras2020}. These techniques significantly enhanced model convergence and stability. This ensured reliable classification under varying conditions. Extensive data preprocessing, including image augmentation and normalization\cite{Elmetwally2024},\cite{Szegedy2015} has proven crucial in mitigating the effects of variations in lighting, background complexity, and obstruction challenges often encountered in real-world surveillance environments. The comparative analysis with traditional object detection models such as YOLOv3\cite{Liu2024} and region-based CNNs\cite{Zhang2016},\cite{Parkhi2015} further demonstrates the superiority of the proposed approach for handling diverse and dynamic visual inputs. Unlike conventional methods which often struggle with fine grained distinctions between intruders and authorized personnel. In contrast, the deep learning-based classifier exhibits higher sensitivity and precision which makes it well-suited for anomaly detection in security essential applications. Beyond accuracy the interpretability remains a crucial aspect of real-world deployment. The integration of SHAP interpretability methods\cite{Breiman2001} and meaningful perturbation analysis\cite{Collobert2008} provides deeper insights into the decision making process which can enhance model transparency and trustworthiness. This is particularly important for high-security environments where advanced AI  techniques can ensure that classification results can be verified and understood. Additionally, benchmarking against classical machine learning approaches including support vector networks\cite{Huang2017} and random forests \cite{Ioffe2015} reinforces the advantages of deep learning-based anomaly detection in terms of adaptability, feature learning, and real-time responsiveness. By integrating optimized deep learning models with extensive preprocessing and interpretability techniques this study provides valuable insights into advancing real-time anomaly detection. Future work can explore further enhancements including the incorporation of self-supervised learning techniques, domain adaptation strategies, and real-world deployment testing, to further refine the system’s accuracy and adaptability in dynamic environments.

\section{Conclusion}
This study presents an integrated framework for anomaly detection using computer vision that leverages a TensorFlow-based CNN built on MobileNetV2 along with an OpenCV-driven detection engine. Extensive experimental evaluations demonstrate that the system achieves classification accuracies of 90.20\% for authorized personnel, 98.60\% for intruders, and 75.80\% for non-human scenes, all while maintaining a processing rate of 30 frames per second. The combination of advanced data augmentation and normalization techniques with robust optimization strategies such as Adam optimization\cite{Nag2001} and batch normalization\cite{Elmetwally2024} has proven critical for enhancing model convergence and stability. Furthermore, our approach shows significant advantages over traditional object detection methods in terms of both accuracy and real-time performance. Future work will focus on incorporating interpretability methods and scaling the system to handle larger and more diverse datasets, thereby setting a benchmark for next-generation surveillance systems in anomaly detection.

\section*{Acknowledgment}

The authors gratefully acknowledge the invaluable support and guidance of Prof. Dr. Mohammad Nasir Uddin, P.Eng., whose insights and expertise have greatly contributed to the successful completion of this work. His unwavering encouragement and critical feedback were instrumental throughout the research work.

\bibliographystyle{IEEEtran}  
\nocite{*}                    
\bibliography{Main_work}         

\end{document}